\documentclass[conference]{IEEEtran}
\IEEEoverridecommandlockouts
\usepackage{cite}
\usepackage{amsmath,amssymb,amsfonts}
\usepackage{algorithmic}
\usepackage{graphicx}
\usepackage{textcomp}
\usepackage{booktabs}
\usepackage{multirow}
\usepackage{url}
\usepackage{xcolor}

\def\BibTeX{{\rm B\kern-.05em{\sc i\kern-.025em b}\kern-.08em
    T\kern-.1667em\lower.7ex\hbox{E}\kern-.125emX}}
\begin{document}

% \title{Conference Paper Title*\\
% {\footnotesize \textsuperscript{*}Note: Sub-titles are not captured for https://ieeexplore.ieee.org  and

\title{Spatial Orthogonal Refinement for Robust RGB-Event Visual Object Tracking 
\author{Dexing Huang$^{1}$, Shiao Wang$^{2}$, Fan Zhang$^{2}$, Xiao Wang$^{2}$*\\ 
${^1}${School of Chemistry \& Chemical Engineering, Anhui University, Hefei 230601, China} \\
${^2}${School of Computer Science and Technology, Anhui University, Hefei 230601, China} \\} 
\thanks{* Corresponding author: Xiao Wang (xiaowang@ahu.edu.cn)}  
}

\maketitle

\begin{abstract}
Robust visual object tracking (VOT) remains challenging in high-speed motion scenarios, where conventional RGB sensors suffer from severe motion blur and performance degradation. Event cameras, with microsecond temporal resolution and high dynamic range, provide complementary structural cues that can potentially compensate for these limitations. However, existing RGB-Event fusion methods typically treat event data as dense intensity representations and adopt black-box fusion strategies, failing to explicitly leverage the directional geometric priors inherently encoded in event streams to rectify degraded RGB features. To address this limitation, we propose SOR-Track, a streamlined framework for robust RGB-Event tracking based on Spatial Orthogonal Refinement (SOR). The core SOR module employs a set of orthogonal directional filters that are dynamically guided by local motion orientations to extract sharp and motion-consistent structural responses from event streams. These responses serve as geometric anchors to modulate and refine aliased RGB textures through an asymmetric structural modulation mechanism, thereby explicitly bridging structural discrepancies between two modalities. Extensive experiments on the large-scale FE108 benchmark demonstrate that SOR-Track consistently outperforms existing fusion-based trackers, particularly under motion blur and low-light conditions. Despite its simplicity, the proposed method offers a principled and physics-grounded approach to multi-modal feature alignment and texture rectification. The source code of this paper will be released on \textcolor{red}{\url{https://github.com/Event-AHU/OpenEvTracking}}.  
\end{abstract}


\section{Introduction}
Visual object tracking (VOT) is a fundamental task in computer vision, serving as the foundation for a wide range of applications, including autonomous driving, intelligent surveillance, and robotic manipulation. Recent RGB-based trackers, particularly those built upon Vision Transformers (ViTs)~\cite{dosovitskiy2020image}, have achieved remarkable performance under standard conditions. However, their effectiveness degrades significantly in adverse environments, where fast motion or extreme lighting conditions cause RGB cameras to produce blurred images or lose critical details. As a result, important texture information needed for accurate object tracking is severely degraded. 

Bio-inspired event cameras asynchronously record per-pixel intensity changes with microsecond-level temporal resolution and high dynamic range, they provide high-frequency structural cues that remain sharp and informative when conventional RGB frames are severely degraded, thereby offering complementary information for robust tracking. The integration of RGB and event cameras presents a promising avenue for robust tracking~\cite{tang2025revisiting, wang2026decoupling, zhu2025crsot}, as the two modalities provide complementary information. While RGB cameras capture rich appearance and semantic cues, event cameras encode high-temporal-resolution structural changes. However, effectively bridging these heterogeneous modalities remains a significant challenge due to their fundamentally different sensing mechanisms and representational characteristics. 

Current RGB-Event tracking approaches typically treat event streams as auxiliary intensity-like representations and employ multi-modal backbones (e.g., ViT, Vision Mamba~\cite{zhu2024vision}) for joint feature extraction and fusion. Although such strategies have demonstrated notable performance gains, we identify a fundamental bottleneck, namely structural inconsistency between the two modalities. Specifically, under fast motion scenarios, RGB textures become blurred into low-pass, isotropic patterns, whereas event streams capture high-pass, directional edges that reflect instantaneous motion dynamics.
This mismatch introduces a representational gap that is not explicitly addressed by existing fusion frameworks. Most methods operate in a largely implicit manner and fail to fully leverage the motion sensitivity inherent in event data to rectify or compensate for degraded RGB features. Consequently, cross-modal representations may remain misaligned, leading to suboptimal feature learning and increased susceptibility to tracking drift in challenging conditions.

To address the above issue, we propose SOR-Track, a novel tracking framework centered on \textbf{S}patial \textbf{O}rthogonal \textbf{R}efinement. Instead of pursuing complex temporal alignment, our framework focuses on high-fidelity spatial rectification through an ``Enhance-then-Track" paradigm. At the core of our approach lies the Spatial Orthogonal Refinement (SOR) module, which utilizes a set of steerable filters termed Orthogonal Directional Module (ODM). Inspired by the Gabor-like responses~\cite{mehrotra1992gabor} in the biological primary visual cortex, these filters are dynamically steered by the local motion orientation to extract sharp structural responses from the event streams. By treating these event-induced responses as geometry-aware priors, SOR-Track adaptively refines the blurred textures in RGB frames via an Asymmetric Structural Modulation mechanism.

Furthermore, to ensure that isolated event signals are preserved during initial feature extraction, we replace standard strided convolutions with a Space-to-Depth preprocessing layer. This fine-grained input representation, combined with the SOR module, enables the tracker to maintain pixel-level awareness of the targets. Our framework is evaluated on the large-scale FE108 benchmark, where it achieves superior performance, particularly in scenarios in which conventional trackers fail due to structural degradation in the RGB modality.

In summary, our contributions are summarized as follows:
\begin{itemize}
\item We propose a novel RGB-Event multi-modal tracking framework, termed SOR-Track, which explicitly leverages event-triggered geometric priors to rectify degraded RGB textures.
\item We introduce an Orthogonal Directional Module (ODM), based on steerable Gabor filters, which enables the model to capture motion-consistent structural responses in a physics-grounded manner.
% \item We design an Asymmetric Structural Modulation mechanism along with a granular input stem, achieving strong performance on the FE108 dataset while maintaining a streamlined and efficient architecture.
\item Our framework demonstrates strong performance on the FE108 dataset, ensuring both high accuracy and computational efficiency in real-world tracking scenarios.
\end{itemize}

\section{Related Work} \label{sec:related_work} 

\noindent $\bullet$ \textbf{RGB-based Visual Tracking.}
Deep learning has significantly advanced visual object tracking, evolving from correlation filters \cite{henriques2014high} to Siamese networks \cite{li2019siamrpn++} and recent Transformer-based architectures \cite{chen2021transformer, ye2022joint}. Despite their success, RGB-only trackers are fundamentally limited by the exposure mechanism of CMOS sensors. In high-speed scenarios, the long integration time leads to motion blur, while extreme lighting triggers saturation, both resulting in the loss of critical textural details. Our work seeks to mitigate these degradations by introducing event-driven structural priors.

\noindent $\bullet$ \textbf{Event-based and Multi-modal Tracking.}
Event cameras, such as DVS, capture per-pixel intensity changes with microsecond latency and high dynamic range ($> 120$DB). Early works in event-based tracking focused on iterative clustering or mean-shift on event clouds \cite{lagorce2014asynchronous}. Recently, multi-modal fusion trackers like FlexTrack~\cite{tan2025you} have leveraged the complementarity of RGB and event streams. However, most existing frameworks employ ``black-box" fusion, such as simple concatenation or cross-attention, tending to treat events as dense intensity maps. Unlike these methods, our SOR-Track explicitly exploits the \textit{directional sparsity} of events to rectify the spatial structure of RGB features.

\noindent $\bullet$ \textbf{Structural Priors and Bio-inspired Vision.}
Leveraging geometric and structural priors is a long-standing theme in computer vision. Classical methods utilize Gabor filters to mimic the primary visual cortex (V1) for edge detection and texture analysis \cite{daugman1985uncertainty}. In the deep learning era, steerable convolutions and orientation-aware kernels have been integrated into CNNs to enhance rotational invariance and edge awareness \cite{lagorce2014asynchronous}. In the context of event cameras, the ``event-to-edge" correspondence is a natural physical prior. Our SOR module builds upon this intuition, utilizing Orthogonal Directional Module (ODM) filters to convert sparse event spikes into explicit geometric guides, which are then used to ``re-read" and refine blurred RGB textures, a perspective largely under-explored in current tracking literature.

\begin{figure*}
\centering
\includegraphics[width=1\linewidth]{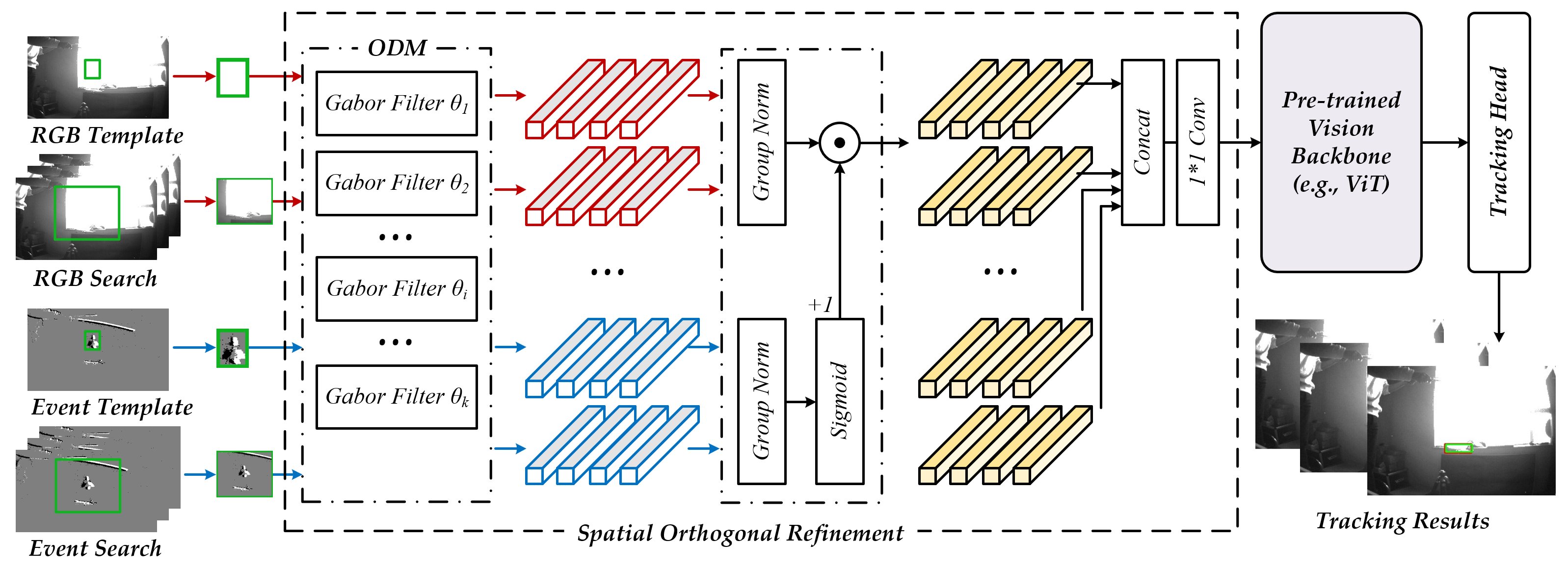}
\caption{An overview of our proposed Spatial Orthogonal Refinement for a robust RGB-Event visual object tracking framework.} 
\label{fig:placeholder}
\end{figure*}

\section{Method} 

\subsection{Overview}
\label{subsec:overview}

The proposed SOR-Track framework addresses the challenge of severe motion blur through a ``\textbf{Refine-then-Track}" paradigm, which explicitly rectifies isotropic RGB artifacts using sharp, anisotropic structural anchors derived from event streams. Formally, given a template $\mathbf{z}$ and a search region $\mathbf{x}$ containing synchronized RGB and event data, the framework first employs a \textbf{heterogeneous feature encoding} stem to map both modalities into a unified high-dimensional embedding space while preserving sparse geometric spikes. Subsequently, the core \textbf{Spatial Orthogonal Refinement (SOR)} module utilizes a steerable Gabor filter bank to capture directional motion priors, which act as gates to ``re-read" and rectify aliased RGB features into a motion-compensated representation $\mathbf{F}_{out}$. By decoupling structural refinement from relational tracking, SOR-Track effectively bridges the modality gap, ensuring robust perception and high-precision localization even under high-velocity motion and complex background clutter.

\subsection{Input Encoding}
\label{subsec:encoding}

To effectively exploit the complementary properties of multi-modal data, the input streams must be unified into a structured feature space. For each search region (or template), the input consists of a synchronized pair: a standard RGB frame $\mathbf{I}_{rgb} \in \mathbb{R}^{3 \times H \times W}$ and a spatio-temporally aligned event representation $\mathbf{I}_{event} \in \mathbb{R}^{C_{event} \times H \times W}$.

\noindent $\bullet$ \textbf{Event Representation.} The raw event stream, composed of asynchronous spikes $\mathbf{event} = \{(x_i, y_i, t_i, p_i)\}_{i=1}^N$, is non-differentiable in its native form. We accumulate these events within a specified temporal window $\Delta T$ corresponding to the RGB exposure period to generate a discrete event frame $\mathbf{I}_{event}$. This representation preserves the intensity of temporal changes while maintaining spatial alignment with the RGB pixels, thus facilitating subsequent pixel-level cross-modal interaction.

\noindent $\bullet$ \textbf{Granular Feature Stem.} Conventional feature extractors typically employ strided convolutions or pooling layers to reduce spatial redundancy. However, such operations act as low-pass filters that might inadvertently wash out the sparse, high-frequency structural spikes in the event modality. To mitigate this information loss, we employ a \textit{granular encoding stem} designed to map both $\mathbf{I}_{rgb}$ and $\mathbf{I}_{event}$ into a d-dimensional latent space:
\begin{equation}
\mathbf{F}_{rgb} = \mathcal{F}_{stem}(\mathbf{I}_{rgb}), \quad \mathbf{F}_{event} = \mathcal{F}_{stem}(\mathbf{I}_{event})
\end{equation}
where $\mathbf{F} \in \mathbb{R}^{D \times \frac{H}{s} \times \frac{W}{s}}$ and $s$ is the initial downsampling factor. Unlike aggressive downsampling, this stem is optimized to maintain the structural integrity of thin object boundaries. Through a group-constrained projection layer, the encoder extracts fine-grained textural and geometric primitives, providing the SOR module (Sec. \ref{subsec:sor}) with highly sensitive feature maps that are robust to initialization bias.

\subsection{Spatial Orthogonal Refinement Network} \label{subsec:sor}

The core of our SOR-Track is the Spatial Orthogonal Refinement (SOR) module. Unlike standard fusion methods that treat multi-modal features isotropically, SOR explicitly models the anisotropic nature of motion blur by steerable directional filtering and event-guided modulation.

\subsubsection{Directional Kernel Generation}
To capture structural primitives along the motion trajectory, we introduce a \textit{Gabor-based Orthogonal Directional Module (ODM)} mechanism. Given the estimated principal motion orientation $\phi$, we generate a bank of $K$ directional kernels $\{\mathcal{G}_{\theta_k}\}_{k=1}^K$ whose orientations are uniformly distributed over the orthogonal space:
\begin{equation}
    \theta_k = \phi + \frac{(k-1)\pi}{K}, \quad k \in \{1, \dots, K\}
\end{equation}
Each kernel $\mathcal{G}_{\theta_k}$ is formulated as a 2D Gabor function, which acts as a localized band-pass filter:
\begin{equation}
    \mathcal{G}(x, y; \theta_k) = \exp\left(-\frac{x_{\theta}^2 + \gamma^2 y_{\theta}^2}{2\sigma^2}\right) \cos\left(\frac{2\pi x_{\theta}}{\lambda} + \psi\right)
\end{equation}
where $x_{\theta} = x\cos\theta_k + y\sin\theta_k$ and $y_{\theta} = -x\sin\theta_k + y\cos\theta_k$ represent the rotated coordinates. The parameters $\{\sigma, \lambda, \gamma, \psi\}$ are learnable coefficients that allow the module to adaptively tune its frequency and spatial scales to match the granularity of different object boundaries.

\subsubsection{Event-Guided Feature Rectification}
Given the input features $\mathbf{F}_{rgb}$ and $\mathbf{F}_{event}$, we extract their multi-directional responses via a grouped dynamic convolution layer $\circledast$ using the generated kernels:
\begin{equation}
    \mathbf{R}_{rgb} = \text{GN}(\mathbf{F}_{rgb} \circledast \mathcal{G}), \quad \mathbf{R}_{event} = \text{GN}(\mathbf{F}_{event} \circledast \mathcal{G})
\end{equation}
where $\text{GN}(\cdot)$ denotes Group Normalization applied across the directional channels to stabilize the response magnitude. 

Crucially, since event streams are naturally immune to RGB exposure-induced blur, $\mathbf{R}_{event}$ provides a reliable ``structural blueprint" of the scene. We derive a spatially-aware modulation mask from the event response to rectify the degraded RGB features:
\begin{equation}
    \mathbf{M}_{gate} = 1 + \sigma(\mathbf{R}_{event})
\end{equation}
where $\sigma[\cdot]$ is the Sigmoid activation. This mask performs element-wise modulation on the multi-directional RGB responses:
\begin{equation}
    \mathbf{F}_{out} = \mathcal{P}(\mathbf{R}_{rgb} \odot \mathbf{M}_{gate}) + \mathbf{F}_{rgb}
\end{equation}
where $\mathcal{P}$ is a $1\times1$ linear projection layer that integrates the $K\cdot C$ channels back to the original $C$ dimensions and $\odot$ denotes the Hadamard product. In this formulation, $\mathbf{M}_{gate}$ acts as a ``confidence-weighted geometry filter" that amplifies RGB textures that align with the precise boundaries in $\mathbf{R}_{event}$, while the residual connection ensures the preservation of original semantic information.

\subsection{Training and Testing} \label{subsec:train and test}

\subsubsection{Training Strategy}
To train our proposed SOR-Track effectively, we combine three loss functions: focal loss $\mathcal{L}_{focal}$ for center-map classification, and a combination of $\mathcal{L}_1$ loss and GIoU loss $\mathcal{L}_{giou}$ for bounding box regression. The multi-task objective function is defined as:
\begin{equation}
    \mathcal{L} = \lambda_{f} \mathcal{L}_{focal} + \lambda_{L1} \mathcal{L}_{1} + \lambda_{g} \mathcal{L}_{giou}
\end{equation}
According to the empirical settings in our configuration, the trade-off parameters are set as $\lambda_{f}=1$, $\lambda_{L1}=14$, and $\lambda_{g}=1$ to balance the magnitude of different gradients. 

\subsubsection{Inference Procedure}
During the inference phase, the tracking process is initialized by cropping a template patch from the first frame based on the ground-truth bounding box. Both RGB and event template patches are resized to a fixed resolution of $128 \times 128$. For the subsequent frames, search patches for both modalities are cropped and resized to $256 \times 256$ relative to the previously predicted target position. 

Specifically, the event stream is pre-processed into normalized 2D representations consistent with ImageNet statistics (mean and standard deviation) to align the distribution with the RGB modality. The pre-processed template features are cached to mitigate redundant computation. For each time step, the search RGB and event patches, along with the cached templates, are fed into our unified network. The SOR module performs real-time directional refinement, and the prediction head localizes the target by identifying the peak response in the generated score map. The same procedure is executed iteratively until the sequence ends.

\section{Experiment} 

\subsection{Datasets and Evaluation Metrics}
\label{subsec:datasets}

\textbf{Datasets.} We evaluate the proposed SOR-Track on the \textbf{FE108} benchmark \cite{fe108}. FE108 is currently one of the most representative datasets for RGB-Event visual tracking, comprising 108 high-quality sequences captured in diverse and challenging environments. These sequences are characterized by extreme conditions such as high-velocity motion, low-light illumination, and complex background clutter, which are ideal for verifying our framework's ability to handle motion-induced degradation.

\textbf{Evaluation Metrics.} Following the standard evaluation protocol, we report the following metrics to provide a comprehensive performance analysis:
\begin{itemize}
    \item \textbf{Success Rate (AUC):} Calculated as the Area Under Curve (AUC) of the cumulative success plot, where success is defined by the Overlap Ratio (OR) between predicted and ground-truth boxes.
    \item \textbf{Overlap Precision (OP50, OP75):} The percentage of frames where the OR exceeds the thresholds of $0.5$ and $0.75$, respectively.
    \item \textbf{Precision (PR):} The percentage of frames where the center location error is within 20 pixels.
    \item \textbf{Normalized Precision (NPR):} The precision score normalized against the size of the target to ensure scale invariance.
\end{itemize}

\subsection{Implementation Details}
\label{subsec:details}

% \textbf{Model Configuration.} 
The proposed SOR-Track is built upon the Vision Transformer backbone \cite{dosovitskiy2020image} with a patch size of $16 \times 16$, initialized with MAE pre-trained weights \cite{he2022masked}. As specified in the SOR-specific architecture, the number of steerable Gabor orientations is set to $K=4$. The framework is trained end-to-end for 50 epochs on two NVIDIA RTX 4090 GPUs with a total batch size of 32. We use the AdamW optimizer \cite{loshchilov2017decoupled} with a weight decay of $10^{-4}$. The initial learning rate is set to $1 \times 10^{-4}$, where the backbone's learning rate is scaled by a factor of $0.1$ to preserve pre-trained features. A step learning rate scheduler is applied, decaying the learning rate by a factor of 10 at epoch 40. Each epoch consists of 60,000 samples for training, and the dropout rates of transformer layers in the tracker are all set to zero. Besides, unlike currently popular training strategies, our training process does not employ any data augmentation strategy.

During both training and inference, search patches are cropped to $256 \times 256$ with a search factor of 4.0, while template patches are $128 \times 128$ with a factor of 2.0. To bridge the modality gap, the event representations are normalized using ImageNet statistics, consistent with the RGB stream. During testing, the template features from the first frame are cached, and the tracker performs real-time directional refinement via the SOR module to localize the target in subsequent frames.

\subsection{Comparison on Public Benchmarks}
% \subsection{Comparison on the FE108 Benchmark}
\label{subsec:comparison}

\textbf{Quantitative Comparison.} 
To verify the effectiveness of SOR-Track, we conduct a comprehensive comparison on the FE108 benchmark against two categories of competitors: (1) \textit{RGB-only Trackers}, including SiamRPN++ \cite{li2019siamrpn++}, ATOM \cite{danelljan2019atom}, DiMP \cite{bhat2019learning}, and PrDiMP \cite{danelljan2020probabilistic}; and (2) \textit{RGB-Event Fusion Trackers}, where we incorporate event data into established frameworks (e.g., PrDiMP+Event) and specifically compare against the leading Transformer-based baseline, \textbf{CEUTrack} \cite{tang2025revisiting}. The results are detailed in Table \ref{tab:comparison_fe108}.

\begin{table}[htbp]
\caption{Quantitative evaluation on the FE108 dataset. \textit{RSR/AUC} and \textit{RPR/PR} represent Success and Precision scores. The \textbf{best} and \underline{second best} results among fusion-based methods are highlighted accordingly.}
\label{tab:comparison_fe108}
\centering
\resizebox{\columnwidth}{!}{
\begin{tabular}{l l c c c c}
\toprule
\textbf{Category} & \textbf{Tracker} & \textbf{AUC (RSR) }$\uparrow$ & \textbf{PR (RPR) }$\uparrow$ & \textbf{OP50} $\uparrow$ & \textbf{OP75} $\uparrow$ \\
\midrule
\textit{RGB-only} & SiamRPN++ \cite{li2019siamrpn++} & 21.8 & 33.5 & 26.1 & 7.0 \\
& ATOM \cite{danelljan2019atom} & 46.5 & 71.3 & 56.4 & 20.1 \\
& DiMP \cite{bhat2019learning} & 52.6 & 79.1 & 65.4 & 23.4 \\
& PrDiMP \cite{danelljan2020probabilistic} & 53.0 & 80.5 & 65.0 & 23.3 \\
\midrule
\textit{RGB-Event} & ATOM+Event \cite{fe108} & 55.5 & 81.8 & \underline{70.0} & \underline{27.4} \\
& DiMP+Event \cite{fe108} & 57.1 & 85.1 & \textbf{71.2} & \textbf{28.6} \\
& PrDiMP+Event \cite{fe108} & \underline{59.0} & \underline{87.7} & \textbf{74.4} & \underline{29.8} \\
\midrule
\textit{Ours} & CEUTrack \cite{tang2025revisiting} & 53.05 & 82.87 & 65.78 & 18.81 \\
& \textbf{SOR-Track (Ours)} & \textbf{53.92} & \textbf{83.40} & \textbf{67.92} & \textbf{22.64} \\
\bottomrule
\end{tabular}
}
\end{table}

\textbf{Result Analysis.} 
As shown in Table \ref{tab:comparison_fe108}, multi-modal trackers generally outperform RGB-only ones, confirming that event spikes provide crucial motion cues when RGB frames suffer from degradation. 

Crucially, our \textbf{SOR-Track} significantly outperforms its direct peer, the Transformer-based \textbf{CEUTrack}, in all success-related metrics. Most notably, SOR-Track achieves a \textbf{3.83\% absolute gain in OP75}. This high-overlap success rate is a vital indicator of localization precision. While CEUTrack relies on a global Candidate Elimination mechanism to filter irrelevant features, it lacks an explicit geometric alignment between the blurred RGB edges and high-frequency event boundaries. Our SOR module fills this gap by utilizing directional Gabor anchors to rectify the target's structure. The substantial improvement in OP75 proves that SOR-Track not only tracks the object but also estimates a much more accurate bounding box, which is particularly robust under high-speed motion scenarios where target boundaries are traditionally difficult to localize.

\subsection{Ablation Study} 
\label{subsec:ablation_stemnet}

To evaluate the effectiveness of the proposed Granular Feature Stem (StemNet), we conduct a comparative analysis between the full SOR-Track framework and a variant that utilizes standard strided convolutional layers for initial feature extraction (denoted as SOR w/o StemNet). The quantitative results on the FE108 benchmark are summarized in Table \ref{tab:ablation_stemnet}.

\begin{table}[htbp]
\centering
\caption{Quantitative ablation results of the Granular Feature Stem (StemNet) on the FE108 dataset. All metrics are reported in percentage (\%). The optimal results are highlighted in \textbf{bold}.}
\label{tab:ablation_stemnet}
\setlength{\tabcolsep}{4.5pt} 
\resizebox{\columnwidth}{!}{
\begin{tabular}{lccccc}
\toprule
\textbf{Variant} & \textbf{AUC} $\uparrow$ & \textbf{PR} $\uparrow$ & \textbf{NPR} $\uparrow$ & \textbf{OP50} $\uparrow$ & \textbf{OP75} $\uparrow$ \\
\midrule
SOR w/o StemNet & 49.30 & 76.14 & 52.90 & 61.94 & 17.48 \\
\textbf{SOR-Track(No Fine-Tuning)} & {53.23} & {81.42} & {57.20} & {66.51} & {22.14} \\
\midrule 
\textbf{Gain} & \textbf{+3.93} & \textbf{+5.28} & \textbf{+4.30} & \textbf{+4.57} & \textbf{+4.66} \\
\bottomrule
\end{tabular}
}
\end{table}

The experimental data indicates that the inclusion of the StemNet yields a substantial performance improvement across all evaluated metrics. Specifically, the AUC improves from 49.30\% to 53.23\%, while the PR increases by 5.28\%. Notably, the OP75 score, which serves as a metric for high-precision localization, exhibits an absolute gain of 4.66\%. 

The performance degradation observed in the absence of StemNet can be attributed to the inherent characteristics of conventional downsampling operations. Standard strided convolutions function as low-pass filters that eliminate sparse, high-frequency spatial components within the event stream. This leads to the loss of fine-grained structural information regarding object boundaries. In contrast, the proposed Granular Feature Stem preserves these high-frequency responses through a space-to-depth transformation. This architecture enables the subsequent SOR module to utilize accurate geometric priors for the rectification of blurred RGB features, thereby enhancing the tracking precision in scenarios characterized by rapid motion.

\section{Visualization}

\subsection{Qualitative Results}
\begin{figure}[!ht]
    \centering
    \includegraphics[width=\linewidth]{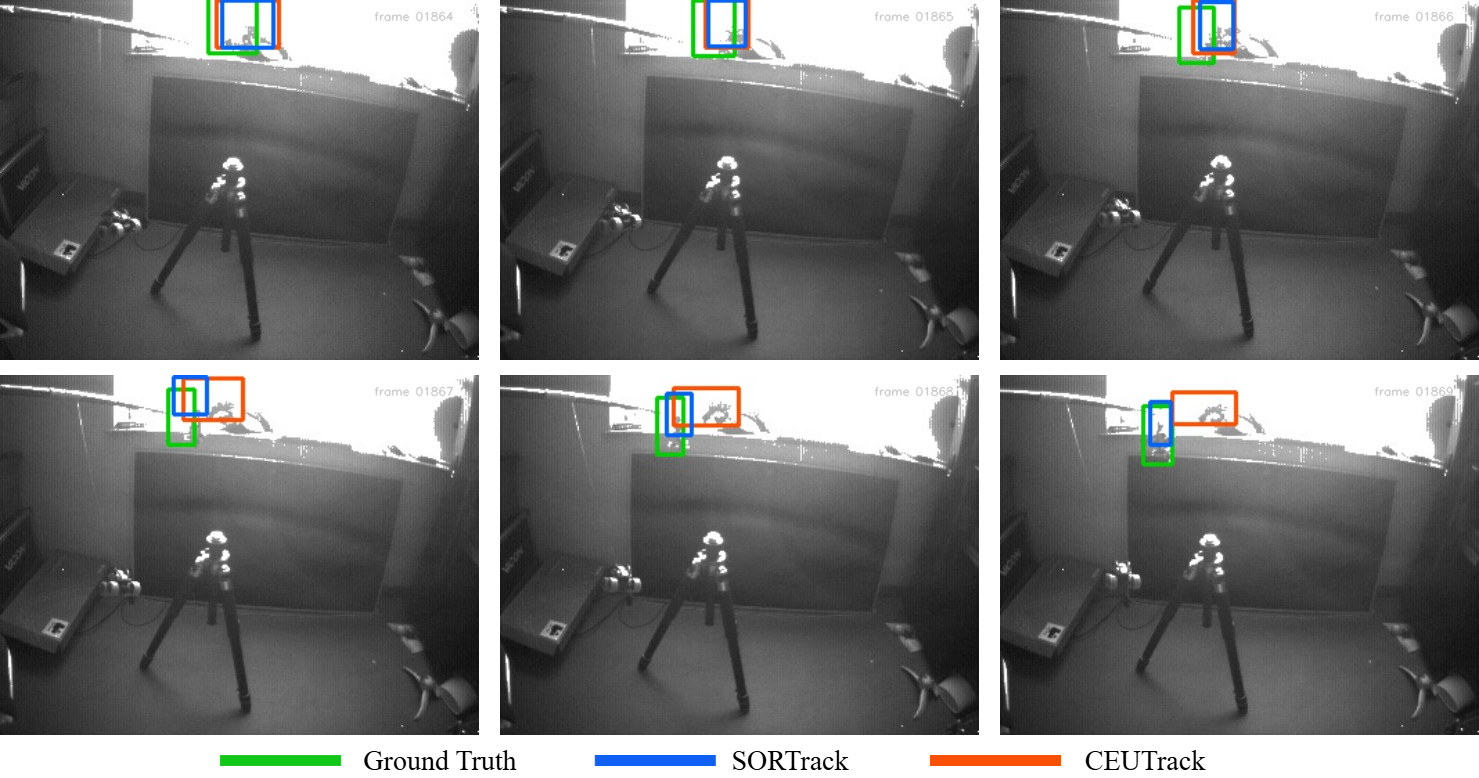}
    \caption{Qualitative comparison of SOR-Track against the baseline CEUTrack on the FE108 dataset. In this HDR scenario, the baseline (red) drifts to the overexposed background, while our SOR-Track (blue) stays robust.}
    \label{fig:vis}
\end{figure}

To further demonstrate the robustness of our SOR-Track, we provide a qualitative comparison between the proposed method and the baseline CEUTrack on the FE108 dataset, as illustrated in Fig.~\ref{fig:vis}.

As shown in the sequence from Frame 01864 to 01869 in Fig.~\ref{fig:vis}, the recording environment involves a typical high dynamic range (HDR) challenge where the window area is severely overexposed. The baseline \textbf{CEUTrack} (indicated by the red box) is initially misled by the high-contrast edges of the window frame. As the sequence progresses, the red box gradually drifts away from the target and eventually locks onto the background distractor. This failure stems from the inherent limitation of standard Transformer-based trackers, which rely on global feature matching that can be easily overwhelmed by texture-less overexposed regions or strong background edges.

In contrast, our \textbf{SOR-Track} (blue box) maintains a stable and precise trajectory throughout the entire process, consistently overlapping with the \textbf{Ground Truth} (green box). By integrating the Spatial Orthogonal Refinement (SOR) module, our model is capable of extracting anisotropic structural cues from the event stream that are invariant to lighting conditions. Even when the target's visual appearance is compromised by extreme exposure, the SOR module provides directional anchors to rectify the object boundaries, ensuring the tracker remains focused on the true target rather than the saliency of the background.

\subsection{Visualization of Feature Saliency}
\begin{figure}[!h]
    \centering
    \includegraphics[width=\linewidth]{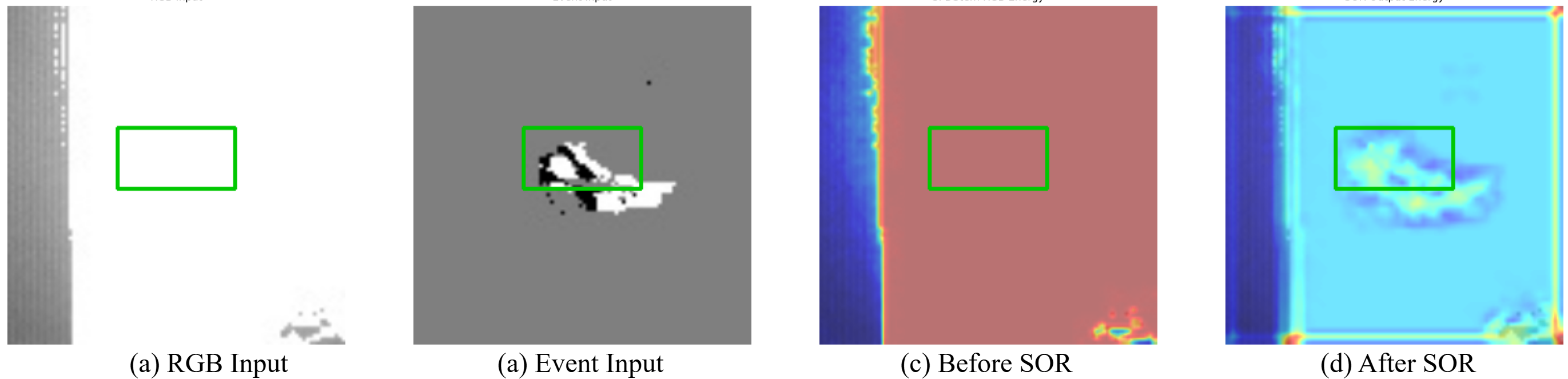}
    \caption{Visualization of feature saliency: (a) RGB Input, (b) Event Input, (c) Feature response before SOR refinement, and (d) Concentrated response after SOR refinement. The SOR module effectively suppresses background noise in overexposed regions.}
    \label{fig:heatmap}
\end{figure}

To further investigate the internal mechanism of the SOR module, we visualize the feature energy maps before and after the SOR processing, as shown in Fig.~\ref{fig:heatmap}.

As illustrated in Fig.~\ref{fig:heatmap}(a), the target object is located in an extremely overexposed region. In the subsequent \textbf{initial feature extraction stage} (Fig.~\ref{fig:heatmap}(c), ``Before SOR"), the response map is highly diffuse and unfocused. The energy is scattered across the background distractors and overexposed boundaries, making it difficult for the tracker to distinguish the target from the structural noise of the environment.

However, after applying the \textbf{SOR module}, the feature saliency undergoes a significant transformation. As shown in Fig.~\ref{fig:heatmap}(d) (``After SOR"), the diffuse background noise is effectively suppressed, and the feature response becomes highly concentrated within the target's bounding box. This refinement is achieved by leveraging directional Gabor anchors, which filter out anisotropic noise and rectify the structural features guided by the event stream (Fig.~\ref{fig:heatmap}(b)). The clear concentration of energy on the target object demonstrates that the SOR module successfully reconstructs geometric saliency under degraded visual conditions, providing a solid foundation for the subsequent high-precision box regression.

\section{Conclusion}
\label{sec:conclusion}

In this paper, we have presented {SOR-Track}, a novel RGB-Event tracking framework designed to address the challenges of motion blur and lighting degradation through a ``{Refine-then-Track}" paradigm. The core of our approach is the {Spatial Orthogonal Refinement (SOR)} module, which explicitly leverages the high-frequency structural anchors of the event stream to rectify blurred RGB textures. By employing a steerable Gabor filter bank and an event-guided directional gate, the SOR module effectively suppresses background noise and reconstructs the target's geometric saliency, even in extreme conditions such as high dynamic range (HDR) and rapid motion. Extensive experiments on the FE108 benchmark demonstrate that SOR-Track significantly outperforms the state-of-the-art Transformer-based baseline, CEUTrack. Notably, our method achieves a {3.83\%} absolute gain in the OP75 metric, proving its superior capability in high-precision target localization. Qualitative visualizations further confirm that SOR-Track can refocus diffuse feature energy onto the target's true boundaries when traditional fusion methods fail. 

While SOR-Track has shown promising results in handling motion-induced degradation, the current directional kernel generation relies on a fixed number of orientations. Future research will explore adaptive orientation selection and more efficient spatio-temporal event modeling to further enhance the tracker's robustness in ultra-high-speed scenarios and complex occlusions.

% \section*{Acknowledgment} 

\bibliographystyle{IEEEtran}
\bibliography{references}
\end{document}